\documentclass[letterpaper, 10 pt, journal, twoside]{IEEEtran} 

\IEEEoverridecommandlockouts 

\usepackage[utf8]{inputenc}
\usepackage{cite}

\usepackage[caption=false]{subfig} 

\usepackage{graphicx} 

\usepackage{amsmath}

\usepackage[binary-units=true,separate-uncertainty=true,multi-part-units=single,per-mode=symbol]{siunitx}
 
\usepackage[dvipsnames]{xcolor}
\newcommand{\revision}[1]{#1}

\title{Where Should I Look? Optimised Gaze Control for Whole-Body Collision Avoidance in Dynamic Environments
}  

\author{Mark Nicholas Finean$^1$, Wolfgang Merkt$^1$, and Ioannis Havoutis$^1$%
\thanks{Manuscript received: September, 9, 2021; Revised November, 17, 2021; Accepted December, 7, 2021.}
\thanks{This paper was recommended for publication by Editor M. Vincze upon evaluation of the Associate Editor and Reviewers' comments. 
This research was supported by UKRI/EPSRC through the University of Oxford Centre for Doctoral Training, Autonomous Intelligent Machines and Systems (AIMS, grant references EP/L015897/1 and EP/R512333/1), and the grants [EP/S002383/1],  [EP/R026084/1] and [EP/R026173/1]. This work was part of the Human-Machine Collaboration Programme, supported by a gift from Amazon Web Services.} 
\thanks{$^{1}$ All authors are with the Oxford Robotics Institute, University of Oxford
        {\tt\footnotesize mfinean@robots.ox.ac.uk}}%
\thanks{Digital Object Identifier (DOI): see top of this page.}
}

\begin{document}
\maketitle
\begin{abstract}
As robots operate in increasingly complex and dynamic environments, fast motion re-planning has become a widely explored area of research. In a real-world deployment, we often lack the ability to fully observe the environment at all times, giving rise to the challenge of determining how to best perceive the environment given a continuously updated motion plan. We provide the first investigation into a `smart' controller for gaze control with the objective of providing effective perception of the environment for obstacle avoidance and motion planning in dynamic and unknown environments. We detail the novel problem of determining the best head camera behaviour for mobile robots when constrained by a trajectory. Furthermore, we propose a greedy optimisation-based solution that uses a combination of voxelised rewards and motion primitives. We demonstrate that our method outperforms the benchmark methods in 2D and 3D environments, in respect of both the ability to explore the local surroundings, as well as in a superior success rate of finding collision-free trajectories -- our method is shown to provide 7.4x better map exploration while consistently achieving a higher success rate for generating collision-free trajectories. We verify our findings on a physical Toyota Human Support Robot (HSR) using a GPU-accelerated perception framework.
\end{abstract}
\begin{IEEEkeywords}
Mobile Manipulation, RGB-D Perception, Collision Avoidance
\end{IEEEkeywords}

\section{Introduction}

\IEEEPARstart{F}{ast} re-planning, especially for robots operating in dynamic environments, is a challenging and active research area \cite{Park2012ITOMP, Li2019, PhilippSSchmitt2019, gpmp2, Kappler2017}. The vast majority of this prior work focuses on instances with a fixed camera, whether external or robot-mounted. While limited in the `field-of-view', or number of sensors, many mobile robots such as the Toyota Human Support Robot (HSR), or humanoid robots, have movable cameras or sensor heads; this presents the unique challenge of determining where the camera should be looking as the robot moves in real-world environments.

In our previous research \cite{Finean2021}, we proposed and demonstrated an integrated perception and motion planning pipeline. We found that collisions and failure cases of the framework primarily occurred due to poor positioning of the head camera during trajectory execution and re-planning --- in other words, the ability of any motion planner to provide collision-free trajectories is limited by the effectiveness of the perception system in continuously updating the environment representation and identifying obstacles. Robots such as the HSR have additional degrees of freedom, such as pan and tilt, to control the position of the head-mounted camera. We found that simple behavioural heuristics for the camera, such as a fixed camera pose or continuous panning, provided unsatisfactory results and merited further investigation. To our knowledge, this work provides the first investigation into active gaze control for motion planning when constrained by a trajectory. We propose a greedy, voxelised cost-based optimisation as an effective solution and show that our method outperforms heuristic approaches. Further verification of our approach is conducted on a physical HSR.

\begin{figure}[tbp]
\centering
\includegraphics[width=0.95\columnwidth]{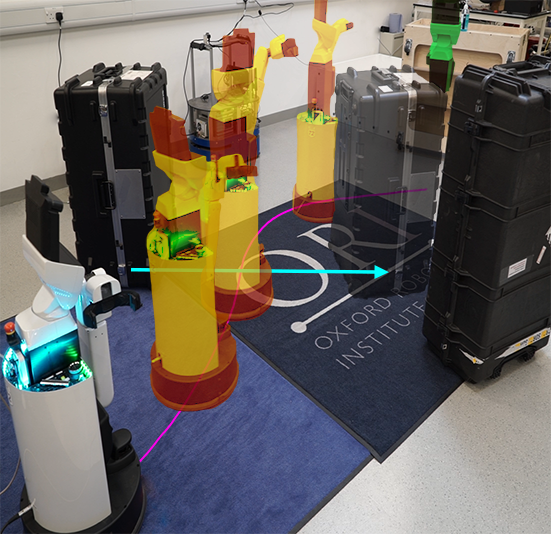}
\label{fig:real_robot}
\caption{Using our method of active gaze control, a Toyota Human Support Robot successfully avoided dynamic obstacles to achieve a goal state (green) from which to pick up a bottle. During execution, a human moved the leftmost large black case into the path of the planned robot trajectory -- the direction of motion is shown with an arrow (cyan). Our optimised gaze control enabled the head-mounted camera to perceive the changes in the environment; the re-planning framework safely and continuously re-planned and executed the task.}
\vspace{-0.5em}
\end{figure}

\subsection{Problem Statement}
\begin{figure}[tbp]
\centering
\includegraphics[width=0.9\columnwidth]{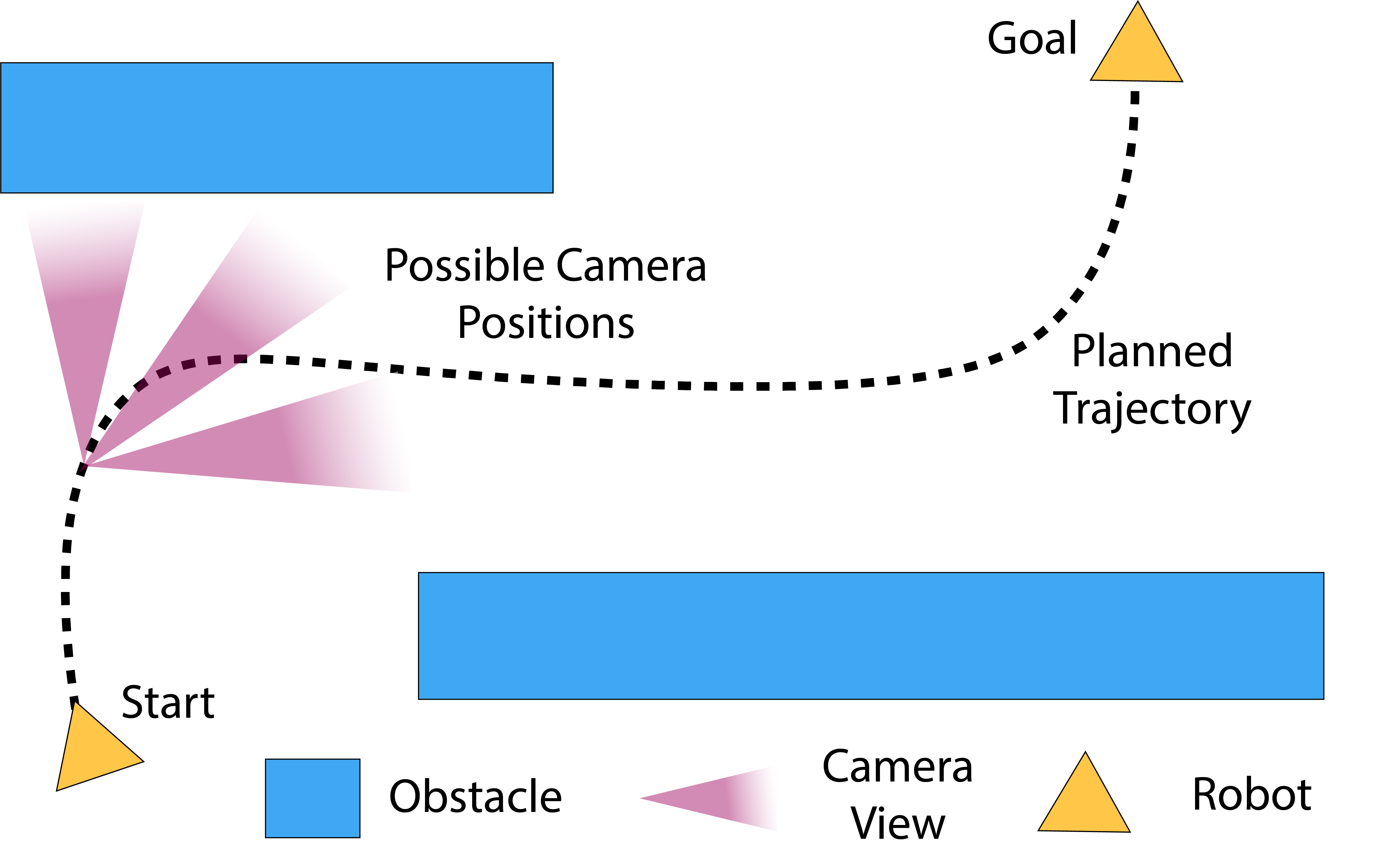}
\caption{At any snapshot in time during a motion planning task, the robot will have partially observed the environment and planned a trajectory accordingly. The problem that we address is in determining where a movable/re-positionable, e.g. head-mounted, camera should be directed in order to provide the most relevant observations for effective collision avoidance and optimised map exploration.}
\label{fig:problem_statement}
\vspace{-0.5em}
\end{figure}

We address the problem of ``given a planned robot trajectory, where should the camera be directed in order to provide both successful collision avoidance and map exploration in potentially dynamic scenes?'' Figure \ref{fig:problem_statement} illustrates the problem. The challenge largely arises because once we start working within the paradigm of dynamic and unknown environments, we must take a more cautious view of our surroundings -- there is no guarantee that an area that was observed in the past will remain static. In particular, for robots operating in environments where humans co-occupy the workspace, there may be the possibility of a human crossing the planned path of the robot; the robot should thus perceive areas of space that it plans to occupy in plenty of time to react to dynamic obstacles.  

\section{Related Work}

Literature on robot camera positioning typically aims to address the `Next-Best View' (NBV) problem whereby the aim is to determine where the camera should be positioned to obtain maximum information gain for the reconstruction of an environment or object model \cite{Potthast2014, Bircher2016, Border, Collander2021, Krainin2011}.

Our problem is subtly different for two reasons. Firstly, in our problem, the camera trajectory is constrained by the planned trajectory of the mobile manipulator on which it is mounted. Secondly, we do not wish to maximise the observance or information gain of an object, or the environment, as in the NBV problem; this in itself does not necessarily assist collision avoidance. Instead, our goal is to optimise our perception of the environment \textit{such that} we achieve resultant trajectories that are collision-free for a given task. In other words, we want our observations of the environment to be relevant to the task and planned robot trajectory. 


We believe that the concept of `trajectory-constrained active gaze control' is a novel problem that has not previously been addressed. A simple approach is to use a fixed head position without any active gaze control. Maier et. al. present and demonstrate an integrated approach for localisation, mapping, and planning in 3D environments using RGB-D cameras \cite{Maier2012}. Using a NAO humanoid robot, they fix the position of the head-mounted depth camera so that the optical axis intersects the floor at a \SI{30}{\degree} angle. The authors found this to be ``the best compromise between observing the near range for obstacle detection and looking ahead for localization and path planning''. While the authors found this to be sufficient, the approach does not fully utilise the ability of the head camera to perceive its surroundings, particularly in a dynamic environment, and is likely to result in collision for holonomic locomotion.

Works that have considered active gaze control have applied it to different problems than what we are concerned with. Lidoris et. al. \cite{Lidoris2007}
present an algorithm to combine trajectory planning and gaze direction control for usage in SLAM. Their objective for gaze control is to minimise estimation errors while exploring an unknown environment.

The work of Seara et. al. is most similar to our problem in determining where to look for obstacle avoidance \cite{Seara2002}. They consider active gaze control for a vision-guided humanoid and determine the pan and tilt of a mounted head camera, however, their objective is to maximise the predicted information gain for the position of \textit{known} objects in the scene. By contrast, our approach focuses on being applicable in unknown environments with 3D collision avoidance.

\section{Greedy Voxelised Cost Optimisation}
Our approach is divided into two parts; the first dealing with reward assignment and the second with the optimization.
\subsection{Reward Assignment}
Our primary concern at all times is that of safety and achieving a collision-free trajectory. Many sampling and search-based motion planners use swept trajectory occupancy to determine valid states for the robot. This assumes that we already know the occupancy of voxels within the swept trajectory. In an unknown, or dynamic environment, this swept trajectory needs to be recently observed by the perception system, e.g. a camera sensor, in order to ensure that it consists solely of free space. If obstacles occupy this space, the motion planner requires this information in order to re-plan for a collision-free trajectory. We propose using a reward for head camera poses that observe swept trajectory regions of the workspace. By observing these regions, the motion planning framework can react and re-plan to avoid collisions with obstacles that are found to be in the way. 

\begin{figure*}[ht]
\subfloat{%
\includegraphics[width=1.0\columnwidth]{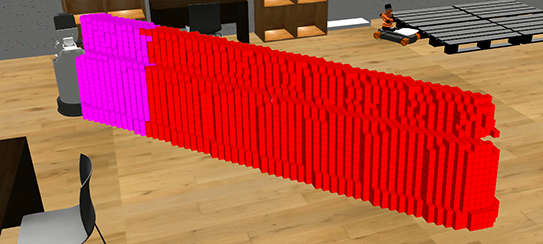}}        
\hfill
\subfloat{%
\includegraphics[width=1.0\columnwidth]{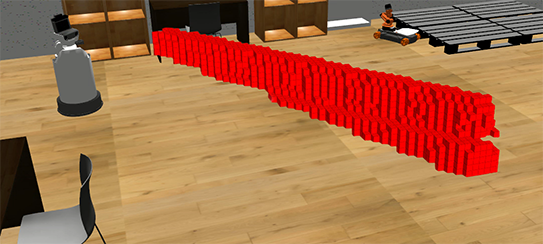}}        
\caption{A visualisation of the swept trajectory rewards as described in our approach to active gaze control. \textit{Left:} A full swept trajectory of rewards that will contribute to the total reward of \revision{field-of-view} cones that contain intersecting voxels. The earlier part of the trajectory (pink) is a higher priority region, with higher reward due to temporal proximity. \textit{Right:} As voxels are observed, their `time last observed' is reset to zero and thus below the threshold for allocating a reward.}
\vspace{-1em}
\label{fig:swept_trajectory}
\end{figure*}

Since trajectories can have large time horizons, we propose a tiered reward that depends on how soon the swept trajectory steps occur in time, thus prioritising the observance of states that occur sooner. This is illustrated in Fig. \ref{fig:swept_trajectory}.

Beyond safety, we wish to explore the environment and maintain a broad perception of the workspace. Being aware of peripheral surroundings is important for enabling the robot to find alternative paths but also for identifying dynamic obstacles early on and incorporating their trajectories into the motion planning. We therefore wish to encourage the robot to have an objective that rewards exploration in its perception strategy. We achieve this by penalising regions of space based on the time that they were last observed. We thus attribute the `time last observed', $t_i$, to each voxel in a maintained voxelmap of the environment. Each time a sensor measurement is received from the camera, the last observed time for all voxels is incremented by one. Voxels within the camera's current field-of-view cone are reset to zero since they have been observed. We thus maintain a voxelmap of observation times.

Formally, we consider a robot trajectory, $\mathbf{X}(t, T)$, that starts execution at time $t$ and has a planned duration of $T$. For such a trajectory, we construct a voxelised map of the swept occupancy, such that the occupancy of a voxel, $v_i$ is:
\begin{equation}
v_i =
    \begin{cases}
      0 & \text{Not occupied}\\
      \tau & \text{Occupied},
    \end{cases}
\end{equation}
where $\tau = [0, T]$ is the earliest time index of the trajectory which results in occupancy.

\revision{%
Using the intuition previously described, we wish to assign reward to regions of space that have not been recently observed. Among these, we want to firstly prioritise observation of the swept robot trajectory, with greater reward assigned to the earlier part of the trajectory and less reward assigned to the later section. Secondly, we wish to provide a reward for broader exploration of the environment. We reflect these three types of reward by assigning each} voxel with a reward, $r_i$, determined by:
\begin{equation}
r_i =
    \begin{cases}
      c_1 & \text{$v_i > 0$ and $v_i <= \tau_s$ and $t_i \revision{>=} \tau_{c}$}\\
      c_2 & \text{$v_i > 0$ and $v_i > \tau_s$ and $t_i >= \tau_{c}$}\\
      max(c_3  t_i, 1) & \text{otherwise}
    \end{cases}
\label{eqn:allocation}
\end{equation}
where $c_1$, $c_2$, and $c_3$ are \revision{positive} constants to be tuned -- these parameters determine the balance between observing the near and far sections of the swept trajectory and exploring the environment. $\tau_s$ represents the farthest step in the currently planned trajectory that is deemed to be a high priority for observation. An example reward allocation for the swept trajectory volume is shown in Fig. \ref{fig:swept_trajectory}. $\tau_{c}$ represents the user-defined threshold for a safe `time last observed' -- a smaller value will result in a more conservative behaviour that is more suited for dynamic environments with fast moving objects. In contrast, a large value is suitable for static environments. \revision{We emphasise that in Equation \ref{eqn:allocation}, $c_1$ and $c_2$ rewards are only allocated if the voxel has not recently been observed, i.e. $t_i >= \tau_{c}$.}

We use the camera's projected cone of vision \revision{(field-of-view cone)} to reset voxel observation times, rather than raycasting, \revision{so as to} avoid reward allocation for regions of space that we know cannot currently be observed. \revision{Raycasting in the perception pipeline is performed between the camera and an observed pointcloud against which the rays will terminate in collision. However, in the gaze control problem, we do not consider whether regions of space are denoted as free, occupied, or unknown---only whether they have been observed recently enough. For example, some camera positions may result in facing a wall; in these cases, a raycasting approach would terminate at the wall (leaving the space behind the surface as `unknown'), and not reset `time last observed' counters on the other side of the wall. The lack of resetting these counters may wrongly encourage the optimisation to face the wall again so as to observe these unobservable voxels.  Hence, we instead opt for a field-of-view cone which instead can result in voxels on the other side of surface boundaries being labelled as observed for the purposes of gaze control optimisation. On first inspection, the reader may be inclined to think that in these instances, the method will now discourage the robot from observing the other side of the wall for future times.} However, this is not a problem due to our method of incrementing the `time last observed', \revision{independent of their occupancy}, and applying a threshold, $\tau_{c}$. In our wall analogy, over time, the robot will try to allocate reward to the previously unobservable areas of space and if the robot has moved to a location in which the robot's vision is no longer obstructed by a wall, then the voxels may be observed in the optimisation.

\subsection{Optimisation}
With the reward of each voxel updated at every timestep, we consider a set of motion primitives local to the camera's current position. For each motion primitive, we consider the `field-of-view' cone and calculate the weighted sum of rewards. We implement a greedy optimisation approach and select the motion primitive with the greatest reward for execution.

When considering a two-dimensional workspace, we found that spatially discretised costs are easily performed using a CPU. However, in three-dimensional workspaces, we leveraged our previous work \cite{Finean2021} by using a GPU-accelerated perception framework based on GPU-Voxels \cite{GPUVoxels, GPUVoxelsMobile}, enabling us to maintain a reactive update frequency for the gaze control. By integrating gaze control with our previous framework, we are able to re-use the same voxelmap information for both whole-body motion planning and gaze control optimisation. In a large $256 \times 256 \times 64$ voxelised workspace, at a resolution of $\SI{5}{\centi \metre}$, we found that our gaze controller operates at \SI{4.9}{\hertz} while optimising over 144 motion primitives -- we note that within each iteration to determine the best camera position, not all 144 motion primitives will necessarily undergo full evaluation as those which violate joint constraints will be terminated early and allocated a maximally negative reward. Higher rates could be achieved by reducing the voxelmap size, reducing the maximum distance of the camera's field-of-view cone, or by considering fewer motion primitives. Hardware specifications are: NVIDIA RTX 2060 GPU, 8-core Intel Core i7-9700 CPU @ \SI{4.50}{\giga\hertz} and \SI{2133}{\mega\hertz} DDR4 RAM.

\section{Heuristic Baseline Methods}

Since, to the best of our knowledge, there does not appear to be a comparable `smart' approach in the literature, we consider three baseline heuristic methods to compare against our proposed optimisation method:
\begin{enumerate}
    \item \revision{Constant}: The camera remains in a fixed position with respect to the base of the robot.
    \item \revision{Panning}: The camera continuously pans at a constant rate throughout the task. The tilt of the camera is fixed such that the optical axis intersects the ground at 30 degrees.
    \item \revision{Look Ahead}: The camera is directed towards the future base position of the robot at a fixed number of time steps ahead in the planned trajectory.
\end{enumerate}

To evaluate the methods, we perform benchmarking on a variety of tasks in both 2D and 3D. In all experiments, the robot is not provided with any knowledge of the environment prior to receiving a goal command, after which all mapping and perception is obtained via live sensor measurements.

\section{Evaluation}
We evaluate our method in static 2D and 3D environments (Sections \ref{sect:2dbenchmarking} and \ref{sect:3dbenchmarking} respectively), followed by evaluation of the methods in a dynamic 3D simulation. Finally, we perform hardware experiments in dynamic scenes (Section \ref{sect:hardware}).

\subsection{2D Benchmarking}
\label{sect:2dbenchmarking}

In 2D, we consider a $1000 \times 1000$ workspace in which the outer perimeter is occupied by walls. Static obstacles are generated randomly within the workspace in which a robot performs motion planning tasks between start and goal state pairs $(x,y,\theta)$.

We generate 150 different environments, comprising of five static rectangular obstacles of random size between 10 and 30 units (cells) in each dimension. We generate 20 different start/goal pairs for evaluation on these environments. A \textit{task} comprises of an environment selection and a start/goal pair. For each task, we implemented each of the four head behaviours in tandem with a GPMP2 motion planner \cite{gpmp2} executed in a receding-horizon manner as detailed in our previous work \cite{Finean2021}. We consider a mounted camera with pan joint values in the range of $[-\frac{\pi}{2}, \frac{\pi}{2}]$, relative to the front of the robot. For the motion primitives in our optimisation, we consider values at $\frac{\pi}{16}$ intervals within this range, resulting in 17 possible head camera positions to greedily optimise. To evaluate the success of each method, we consider both whether the resultant trajectory was collision-free, and the percentage of the environment explored during execution. 
\begin{figure}[ht]
\centering
     \subfloat{%
     \hspace{0.075\columnwidth}
      \includegraphics[width=0.4\columnwidth]{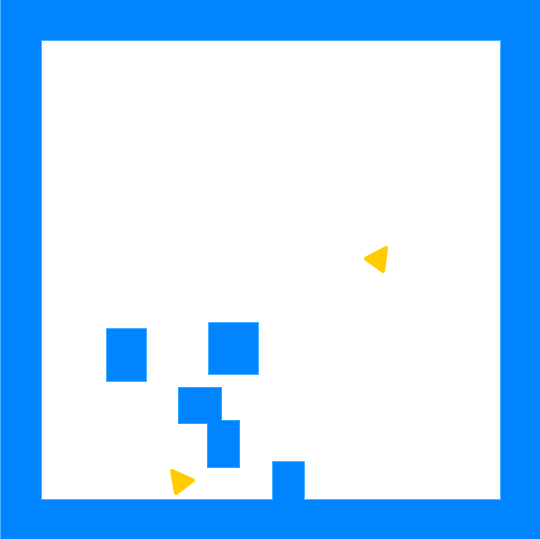}}        
      \hfill
       \subfloat{%
      \includegraphics[width=0.4\columnwidth]{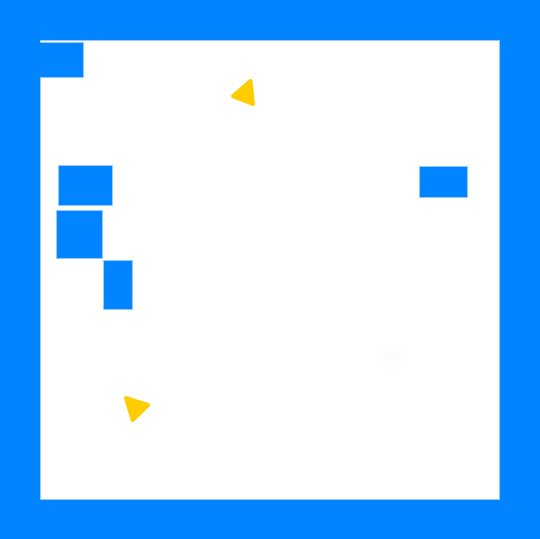}}        
     \hspace{0.075\columnwidth}
      \vfill
    \hspace{0.075\columnwidth}
        \subfloat{%
      \includegraphics[width=0.4\columnwidth]{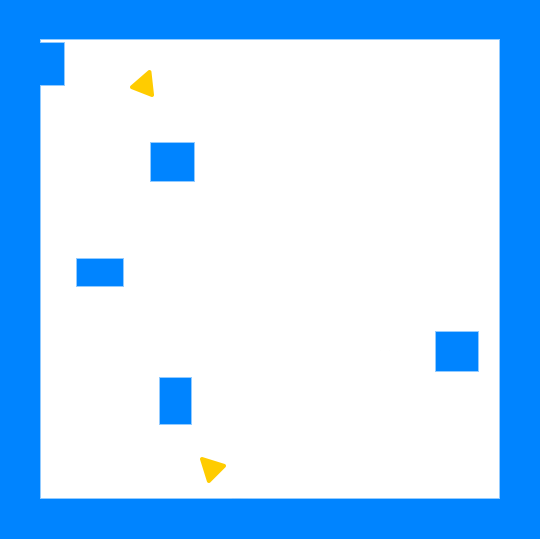}}        
      \hfill
       \subfloat{%
      \includegraphics[width=0.4\columnwidth]{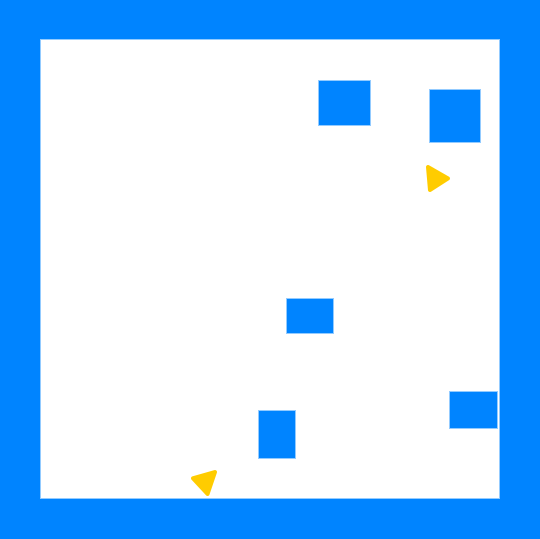}}   
    \hspace{0.075\columnwidth}
\caption{A sample of 2D motion planning tasks comprising of a perimeter wall, obstacles (blue), and a start/goal state pair (yellow triangles).}
\label{fig:2Dexamples}
\end{figure}
A selection of 2D task environments are shown in Fig. \ref{fig:2Dexamples}.

By conducting a parameter sweep for a series of similar tasks, we found the optimum values of $c_1$, $c_2$, and $c_3$, in our setup to be $10^6$, $10^3$, and $1$ respectively. \revision{We found that values of $\tau_s = 3$ and $\tau_{c} = 3$ performed well. For the field-of-view cones, we use an opening angle of $\frac{\pi}{2}$ and a distance cutoff of 200 pixels.}

Of the 3000 tasks in total, we retained a randomly sampled subset of 2000 tasks in which at least one method succeeded in generating a collision-free trajectory -- this provides us with a fair comparison across the methods and provides assurance that the motion planning problem was solvable.

\begin{figure}[t]
\centering
    \subfloat{%
    \includegraphics[width=0.90\columnwidth]{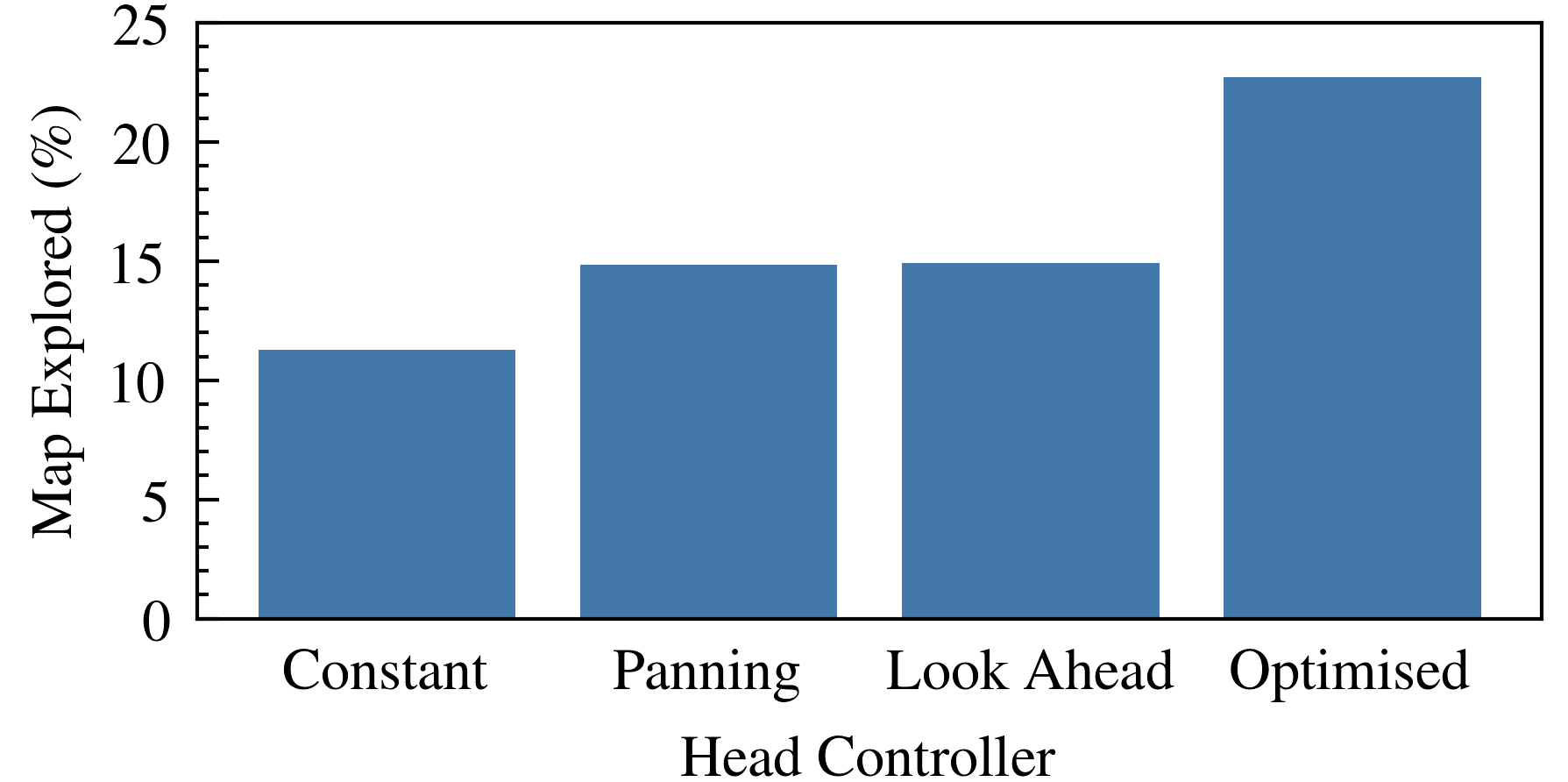}}    
    \vfill
    \subfloat{%
    \includegraphics[width=0.90\columnwidth]{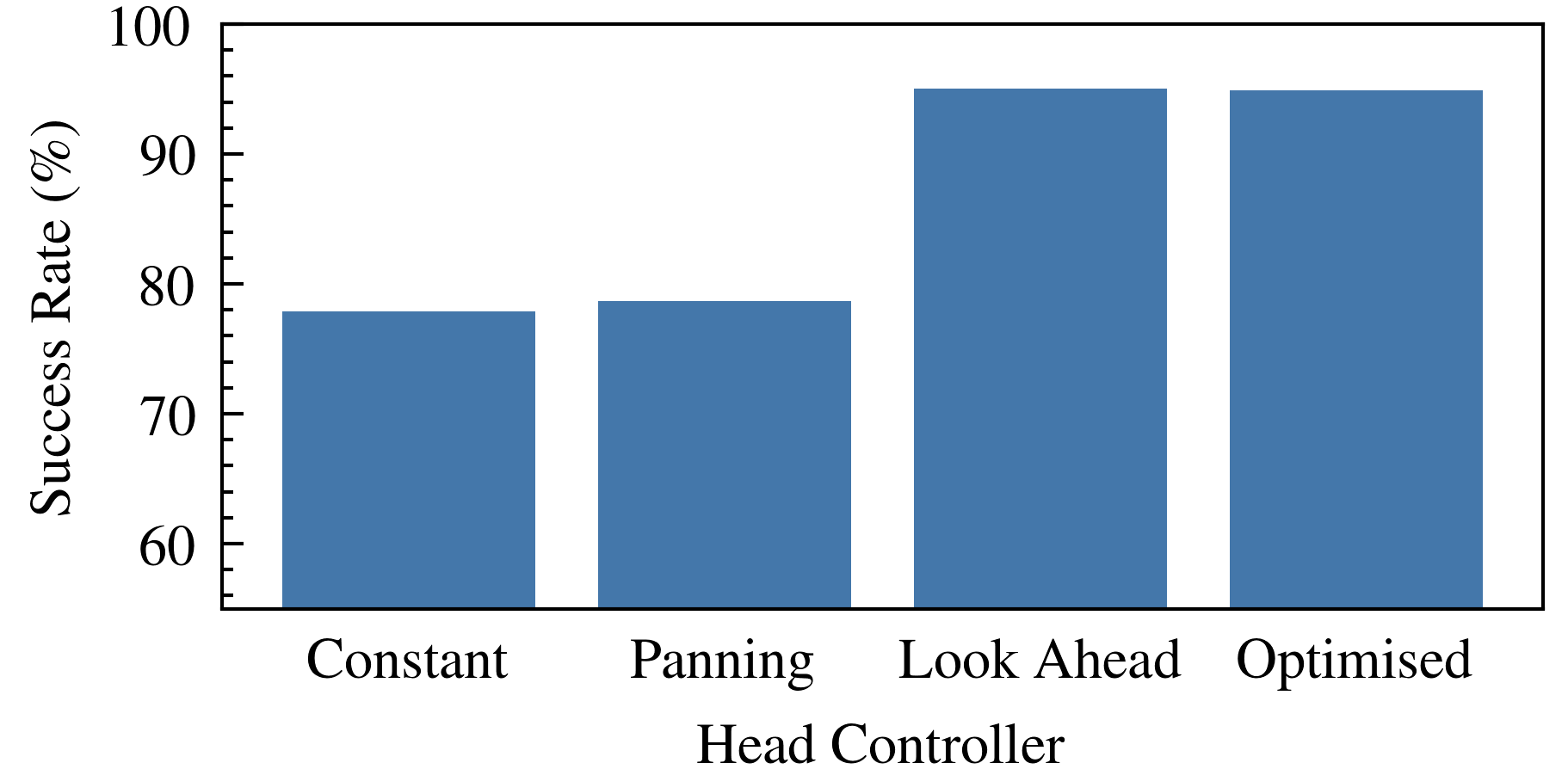}}  
\caption{Top: A comparison of the mean map exploration, as a percentage of the total workspace, achieved by each of the gaze behaviours on our 2D dataset.  \revision{Bottom: A similar comparison of the success rate for each gaze behaviour in achieving collision-free resultant trajectories.} We see that our \textit{Optimised} method provides significantly greater map exploration \revision{while maintaining the joint highest success rate for achieving collision-free trajectories}.}
\label{fig:2Dmapresults}
\vspace{-0.5em}
\end{figure}

\begin{figure}[t]
\centering
\includegraphics[width=1.0\columnwidth]{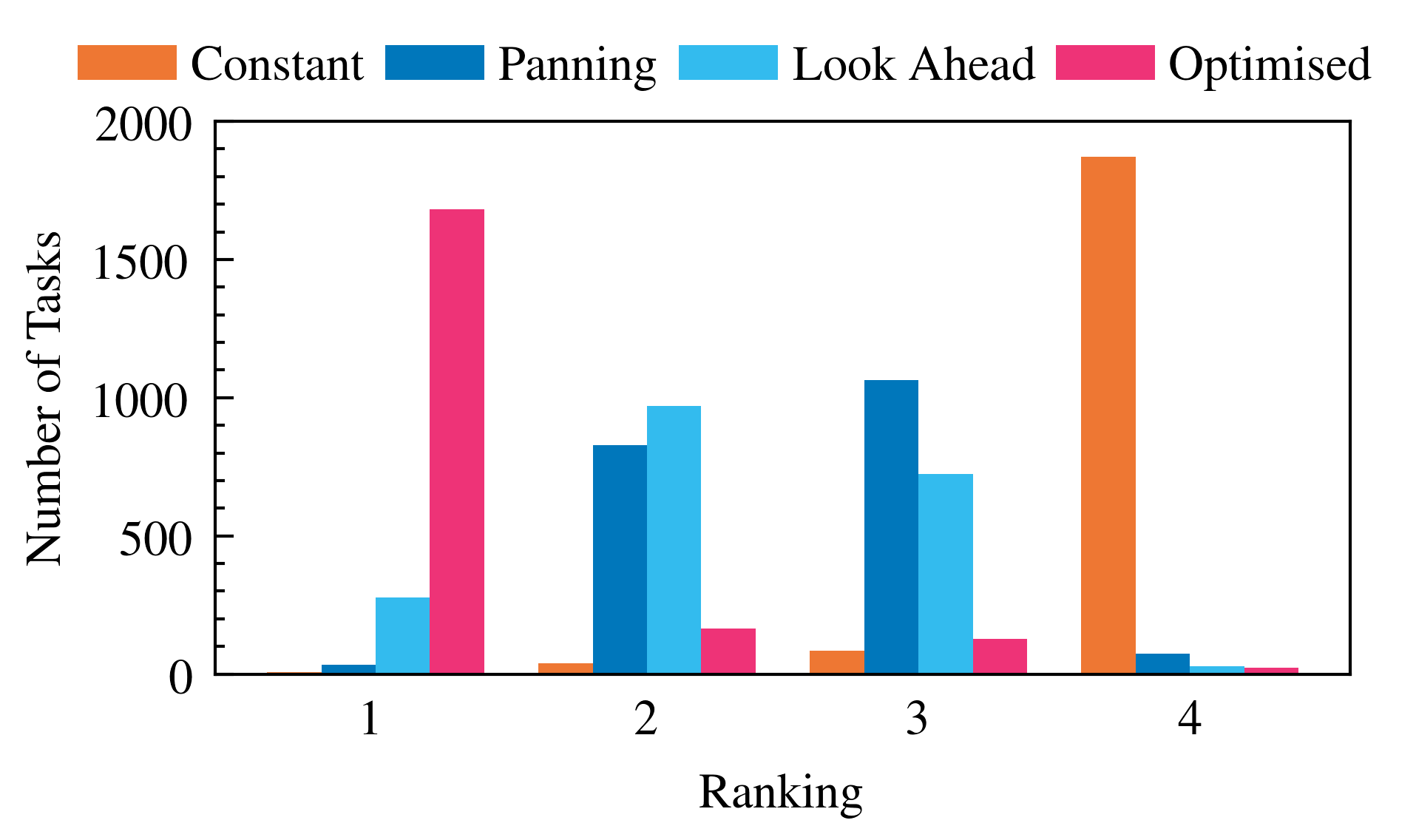}
\vspace{-2em}
\caption{For each task, we rank the gaze control behaviours in a tiered approach, firstly prioritising successful collision avoidance, followed by the map coverage achieved during execution. Across our 2D dataset of tasks, we find that our \textit{Optimised} method outperforms all other methods in 94\% of tasks.}
\label{fig:2Drankingresults}
\vspace{-1em}
\end{figure}

\begin{figure}[ht]
\centering
\includegraphics[width=0.9\columnwidth]{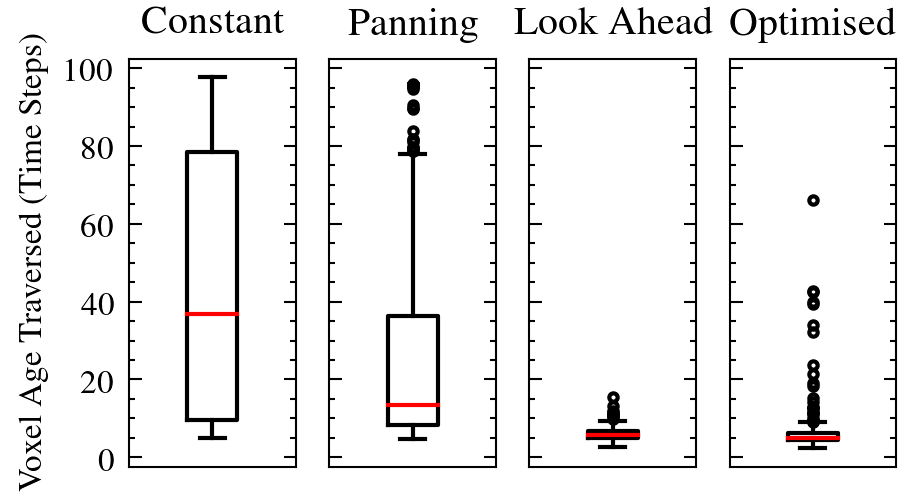}
\caption{Distribution of the last time that cells were observed at the time of robot occupancy. \textit{Last observed times} were initialised and clipped at 100 time steps. Times for voxels that are observed by the camera are reset to zero and incremented for each subsequent time-step that they are not observed. To ensure safe locomotion, cells should be recently observed prior to the robot moving to occupy them. Our \textit{Optimised} method allows us to parameterise our `cautiousness' of the environment and trade-off the conservative behaviour of the \textit{Look Ahead} method, with greater map exploration.}
\label{fig:2DVoxelAge}
\vspace{-1em}
\end{figure}

Results from the 2D benchmarking experiments indicate that our \textit{Optimised} method, along with the \textit{Look Ahead} method, achieve the greatest success rates for producing collision-free trajectories (95\%). This is likely due to maintaining a much better perception of the region of space directly in front of the robot. As shown in Fig. \ref{fig:2DVoxelAge}, these two methods result in the robot traversing cells which have typically been more recently observed. We note again that this feature is of particular interest when considering dynamic environments, since an old observation of a cell may no longer be valid. The additional benefit of our method is a significantly enhanced perception of the surrounding environment due to the exploration reward in the objective function. Figure \ref{fig:2Dmapresults} shows the mean map coverage achieved by each of the four methods over the subset of tasks in which all methods succeeded in finding collision-free trajectories --- our method provides a \SIrange{52}{102}{\percent} improvement over the benchmark approaches in exploring the environment during execution.
As previously discussed, the primary consideration of an effective gaze controller is in providing successful collision avoidance, followed by any additional ability to observe the surroundings. We therefore ranked the four methods hierarchically -- firstly by whether trajectories were collision-free and secondly by the total map coverage. Results of this ranking are shown in Fig. \ref{fig:2Drankingresults} whereby our method ranks first $94\%$ of the time.

\subsection{3D Benchmarking}
\label{sect:3dbenchmarking}

\begin{figure*}[ht]
     \subfloat{%
      \includegraphics[width=0.49\columnwidth]{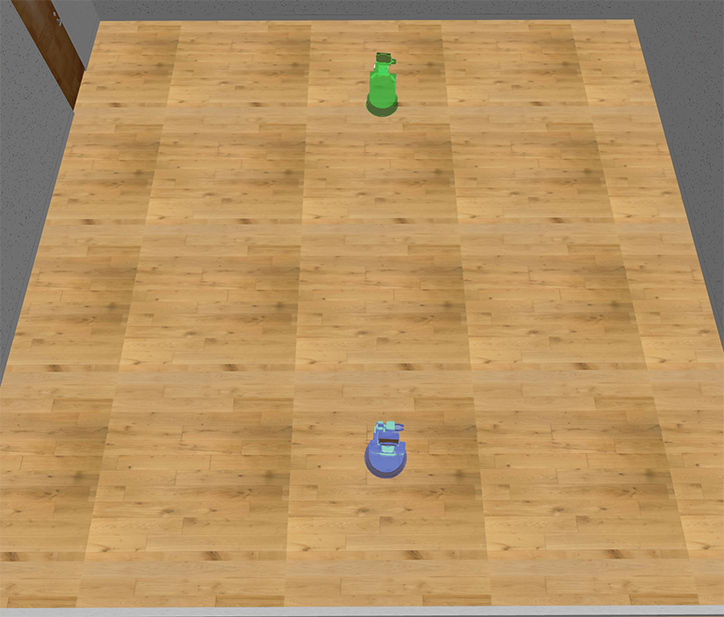}}        
      \hfill
       \subfloat{%
      \includegraphics[width=0.49\columnwidth]{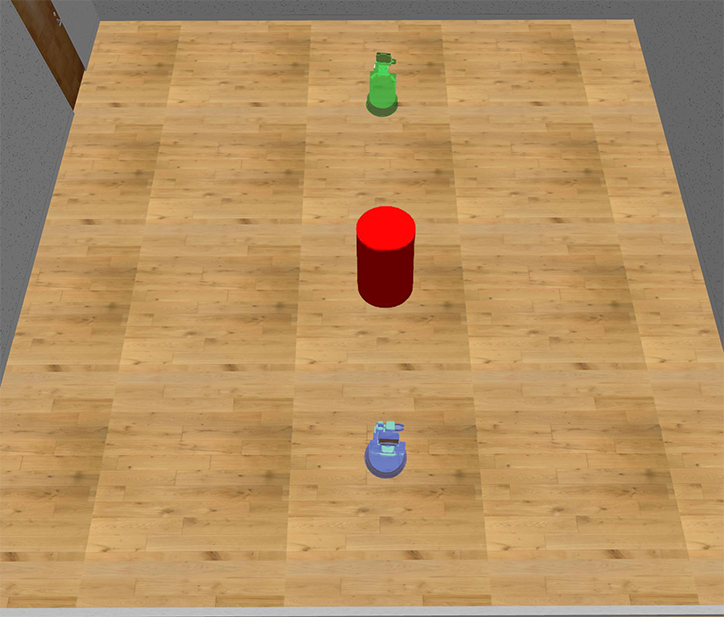}}        
      \hfill
     \subfloat{%
      \includegraphics[width=0.49\columnwidth]{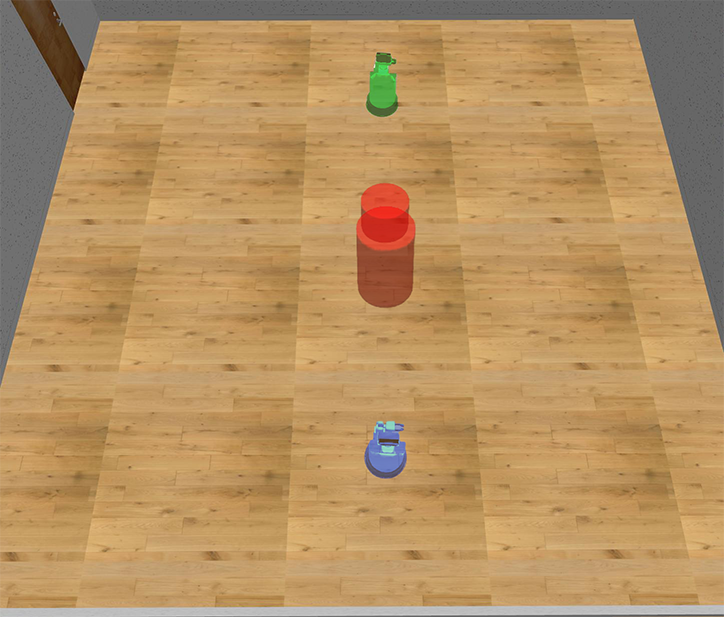}}        
      \hfill
       \subfloat{%
      \includegraphics[width=0.49\columnwidth]{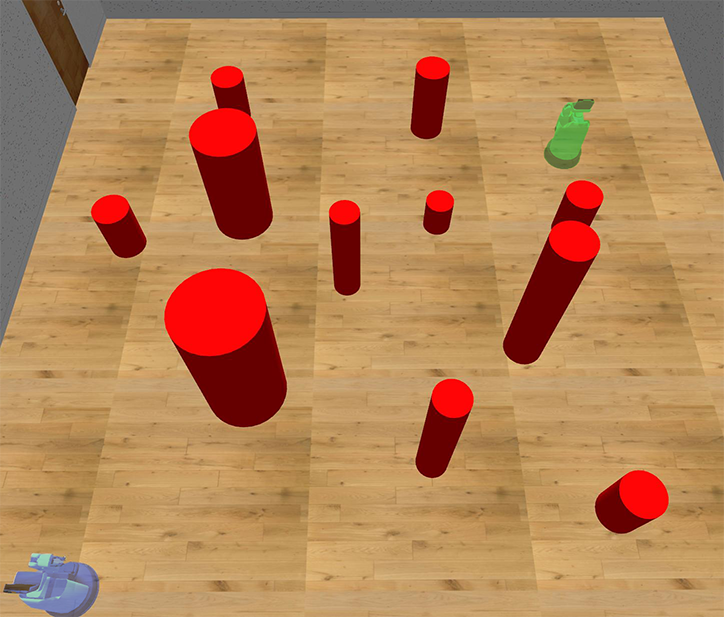}}        
\caption{Aerial view of the four static 3D simulation environments: `Free', `One Obstacle', `Occluded' (a second smaller obstacle is hidden behind the first), and `Clutter'. Start and goal robot states are shown as blue and green respectively.}
\label{fig:3Dexamples}
\end{figure*}

\begin{figure*}[ht]
    \centering
      \subfloat[Constant]{%
      \includegraphics[width=0.49\columnwidth]{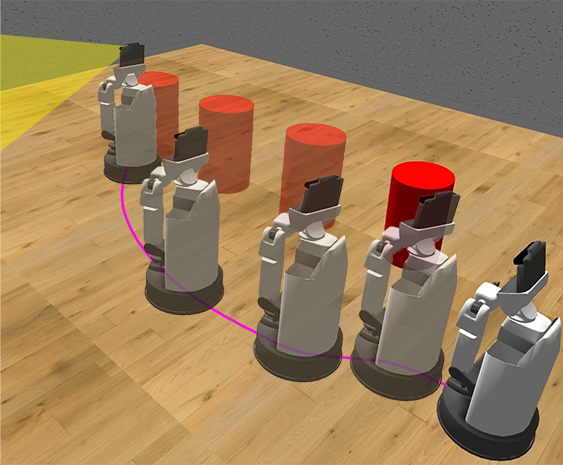}}    
      \hfill
      \subfloat[Panning]{%
      \includegraphics[width=0.49\columnwidth]{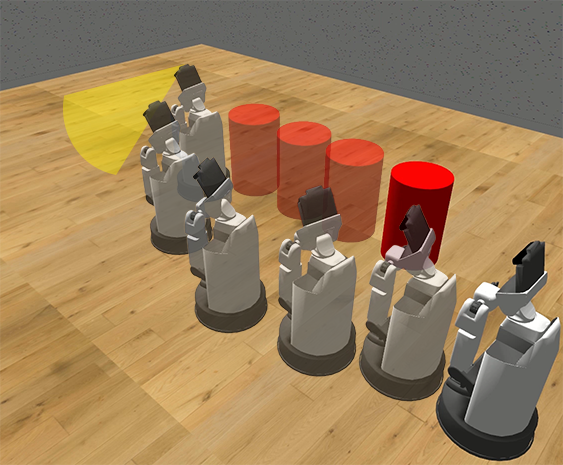}}  
      \hfill
      \subfloat[Look Ahead]{%
      \includegraphics[width=0.49\columnwidth]{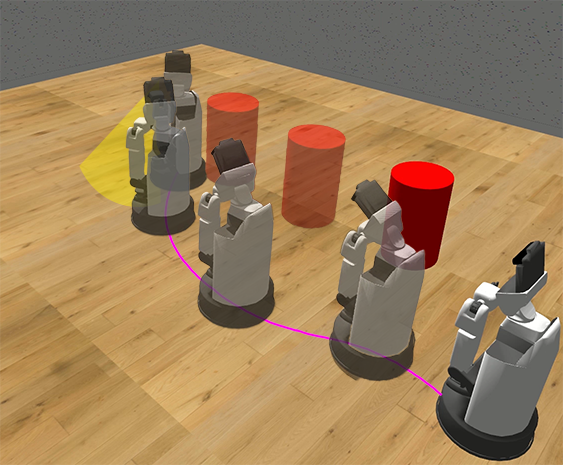}}  
      \hfill
       \subfloat[Optimised (Ours)]{%
      \includegraphics[width=0.49\columnwidth]{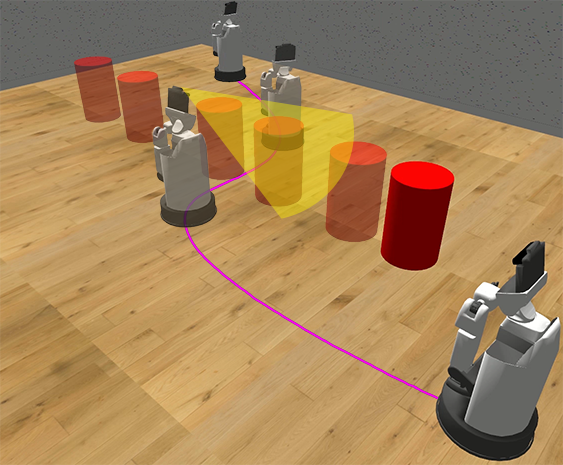}}  
\caption{Our simulated 3D dynamic experiment highlights the benefit of using an active gaze control approach that aims to maintain a broader perception of the environment. In the examples shown, all three benchmark methods demonstrated collisions with the dynamic obstacle. In contrast, our \textit{Optimised} method, which balances the objectives of map exploration and observing the swept volume of the robot's currently planned trajectory, is able to successfully perceive and avoid the dynamic obstacle.}
\label{fig:3DDynamic}
\vspace{-1em}
\end{figure*}

We conducted a series of 3D simulation experiments using a Toyota HSR robot in a variety of static Gazebo environments as shown in Fig. \ref{fig:3Dexamples}, as well as dynamic environment shown in Fig. \ref{fig:3DDynamic}. For each environment, a base goal $(x,y,\theta)$ was provided and our entire pipeline (integrated motion planning, perception, and head controller) was executed so that the HSR would autonomously navigate towards the goal. In each task, after receiving the goal location, the head camera was instructed to first look towards the goal destination and calculate the initial trajectory plan; after which all head movements were controlled by the head controller under investigation.

In our optimisation, we consider 16 joint values for the panning angle and nine values for the tilt, resulting in 144 possible head configurations to optimise over. We recorded whether trajectories were collision-free and the total portion of the map observed during execution. Each task was repeated 10 times for each head behaviour.
\revision{%
In our simulated 3D and hardware experiments, we use a \SI{0.5}{\second} interval between motion planning time-steps and found that a threshold value of $\tau_s = 15$ performed well, corresponding to a time threshold of \SI{7.5}{\second}. We similarly found $\tau_{c} = 150$ voxelmap updates to perform well. For reference, although we continuously re-estimate the trajectory time-horizon during execution, a typical starting horizon for tasks such as in the hardware experiment is $\sim\SI{35}{\second}$. We use a pyramidal field-of-view cone for an ASUS Xtion Pro Live camera, matching the parameters for the sensor characteristics \cite{Swoboda2014}, with horizontal and vertical angles of {58\textdegree} and {45\textdegree} respectively. In line with the sensor's recommended `distance of use', we apply a distance cutoff of \SI{3.5}{\metre}.}

\begin{figure*}[ht]
    \centering
     \subfloat{%
      \includegraphics[width=1.0\columnwidth]{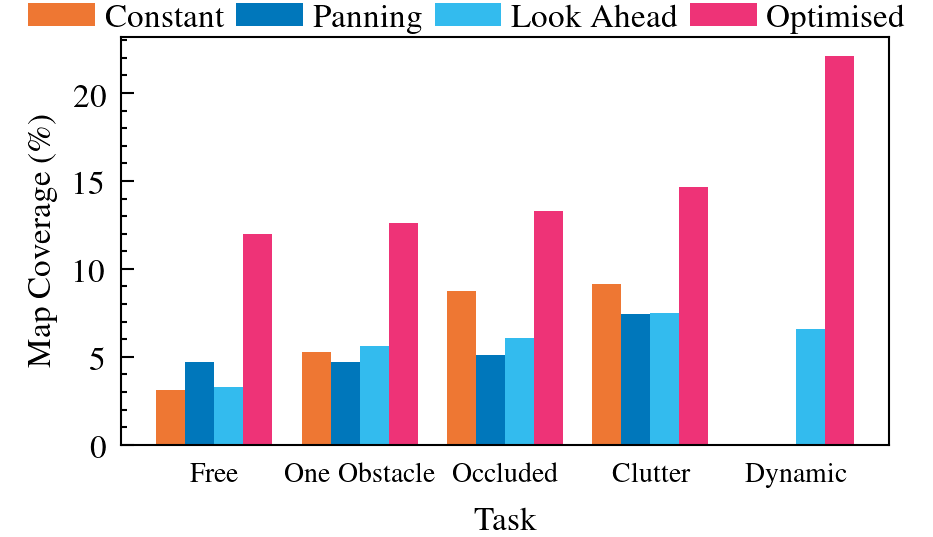}
      \label{fig:3Dmapcoverage}
      }        
      \hfill
     \subfloat{%
      \includegraphics[width=1.0\columnwidth]{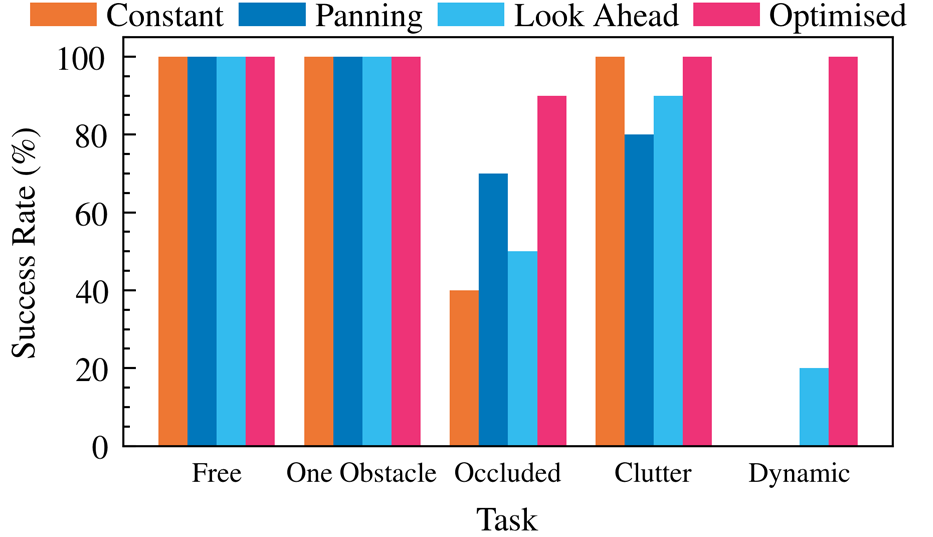}
      \label{fig:3Dsuccessrates}
      }
      \vspace{-0.5em}
\caption{Resulting success rate and mean map coverage achieved by each gaze control method across our 3D simulation tasks. Our \textit{Optimised} method provides consistently high success rates in generating collision-free trajectories, while providing up to 7.4$\times$ improved map exploration in the static environments. Crucially, in the dynamic environment, the \textit{Optimised} approach was the only method to reliably achieve collision-free trajectories.}
\label{fig:3DResults}
\vspace{-1em}
\end{figure*}

Our findings are further validated in the 3D case with results shown in Fig. \ref{fig:3DResults}. In particular, we find that our method is more robust at providing collision-free trajectories and typically provides 2-3x greater map exploration during execution on the static environments, despite taking a similar amount of time. In particular, we note that our proposed optimisation method provides a much higher success rate than the benchmarks on the `Occluded' environment. In this environment, vision of a second smaller obstacle is occluded by the first. The benchmark methods would typically fail to observe the back of the first obstacle as a trajectory around it was executed, thus resulting in collision.

In the 3D dynamic task, the \textit{Constant} and \textit{Panning} methods failed to generate a collision-free trajectory in any of the 10 repeated trials. The \textit{Look Ahead} behaviour found a collision-free trajectory in two of the trials, narrowly passing in front of the obstacle. In contrast, using the \textit{Optimised} approach resulted in a 100\% success rate and achieved significantly greater map coverage, as shown in Fig. \ref{fig:3DResults}.

For the dimensions given previously, the generation of the costmap takes approximately \SI{0.5}{\milli \second}. Cost evaluations for each of the primitives are then performed using this costmap. We found the cost evaluation for a single camera primitive to be \SI{2.4 \pm 0.4}{\milli \second}. In the optimisation, camera primitives that result in a position outside the allowed joint ranges bypass the optimisation and return an infinite (negative) cost. Over an entire trajectory, in which the camera position was determined 160 times, our optimisation approach achieved a mean computation time of \SI{205 \pm 32}{\milli \second} to determine the best camera pose for the next step.

\subsection{Hardware Experiments in Dynamic Environments}
\label{sect:hardware}
\begin{figure*}[t]
          \subfloat[\label{fig:hardware_a}]{%
          \includegraphics[width=0.49\columnwidth]{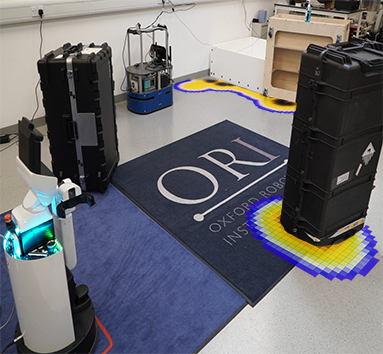}}    
      \hfill
          \subfloat[\label{fig:hardware_b}]{%
          \includegraphics[width=0.49\columnwidth]{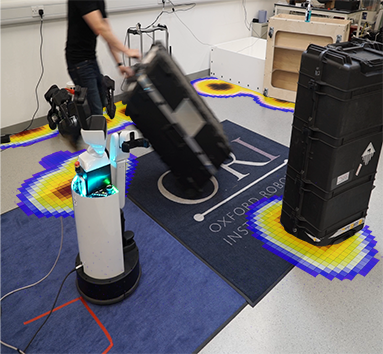}}  
      \hfill
          \subfloat[\label{fig:hardware_c}]{%
          \includegraphics[width=0.49\columnwidth]{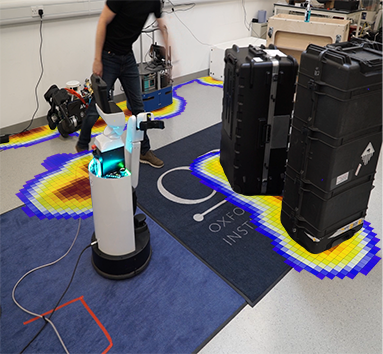}}  
      \hfill
       \subfloat[\label{fig:hardware_d}]{%
        \includegraphics[width=0.49\columnwidth]{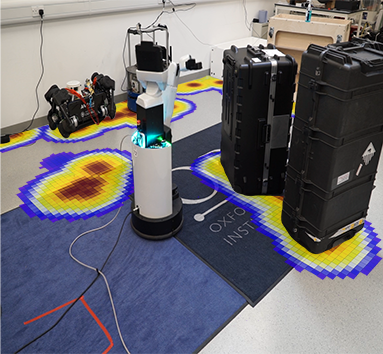}}
      \vfill
       \subfloat[\label{fig:hardware_e}]{%
        \includegraphics[width=0.49\columnwidth]{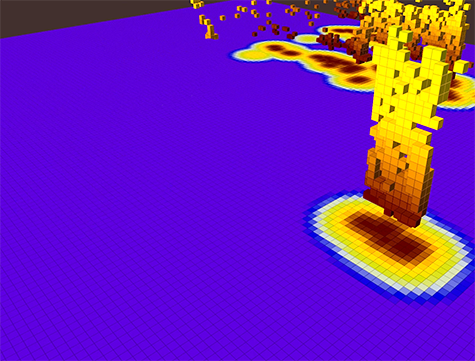}}
      \hfill
       \subfloat[\label{fig:hardware_f}]{%
        \includegraphics[width=0.49\columnwidth]{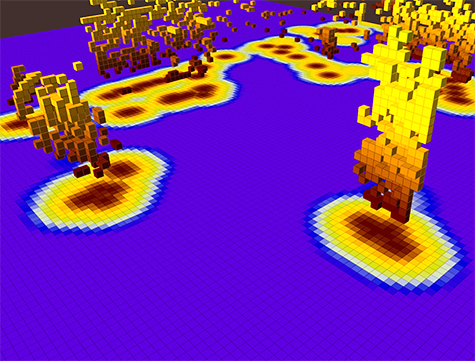}}
      \hfill
       \subfloat[\label{fig:hardware_g}]{%
        \includegraphics[width=0.49\columnwidth]{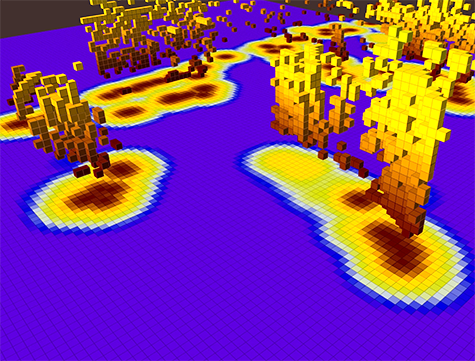}}
      \hfill
       \subfloat[\label{fig:hardware_h}]{%
        \includegraphics[width=0.49\columnwidth]{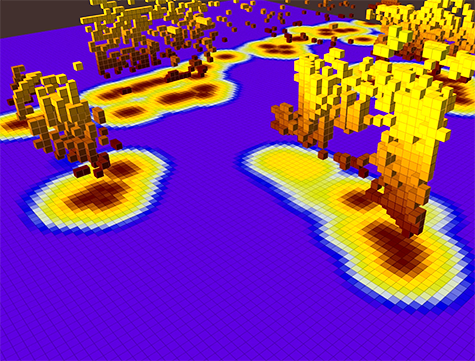}}        
\caption{We validate our proposed method for optimised gaze control on a 8-DoF Toyota Human Support Robot in a live dynamic environment. With time increasing in the images, from left to right, we overlay the calculated ground distance field generated in real-time by the perception system. Between Fig. \ref{fig:hardware_a} and Fig. \ref{fig:hardware_b}, the camera perceives the large obstacle prior to moving. As the dynamic obstacle is moved, the gaze controller continues to balance map exploration with maintaining recent observations of the swept whole-body robot trajectory. In Fig. \ref{fig:hardware_c}, we see that the robot observes the dynamic changes in the environment and is able to quickly re-plan accordingly to avoid collision. Figure \ref{fig:hardware_d} shows the robot taking a collision-free path around the newly observed obstacle.
Note that while we visualise the projected ground distance fields, our method operates using the full 3D voxelmap as shown in the row of images below.}
\label{fig:HardwareSeries}
\vspace{-1em}
\end{figure*}

To demonstrate our approach on hardware, we use a physical Toyota HSR -- an 8-DoF mobile manipulator with a holonomic base and a head-mounted Asus Xtion Live RGB-D camera. The head motion is controlled by a further two degrees of freedom (pan/tilt). The robot is provided with a whole-body goal state to pick up a bottle on the other side of a room. During execution, a human walks into the scene and places a large obstacle in the planned path of the robot. The task is shown in Fig. \ref{fig:real_robot}, while a series of snapshots from this experiment are shown on Fig. \ref{fig:HardwareSeries}.

We found that by using our method, the robot successfully re-observed the space in front of it, perceiving the dynamic change in the environment and successfully re-planning its trajectory to avoid obstacles and pick up the bottle.

Due to the height of the head (and the torso translation joint), this was particularly apparent when the obstacle crossed the path close to the robot; the heuristic methods would typically look \textit{over} the obstacle and fail to re-observe the space immediately in front of the robot's base while the optimised method would alternately observe the space in front of the robot and the last planned robot trajectory. As a result, the observed behaviour of waiting until the path was clear appeared natural.

\section{Discussion}

\revision{Our greedy optimisation approach proves to be a simple solution to the trajectory-constrained gaze control problem that we have described. By only considering the next camera pose, although not meeting real-time requirements of $\sim\SI{30}{\hertz}$, we are able to achieve a sufficiently high update rate of $\sim\SI{5}{\hertz}$. Over longer time-horizons, our method will likely provide sub-optimal solutions that are less smooth than gaze trajectory optimisation performed over a time horizon. Optimising over longer horizons is an interesting topic for further work, however the main concern is potentially large compute times arising from generating and evaluating rewards for subsequent time-steps. This limitation is also shared by the state-of-the-art solutions for the typical next best view problem}.

The benefits of our approach will inevitably vary depending on the specifications of the robot on which it is implemented. For example, robots with a slower, more restricted head motion are likely to see smaller gains in map coverage. Conversely, as hardware improves and systems become more agile, we expect that our approach will lead to further performance gains across a range of platforms.

In examining the trajectory-constrained gaze control challenge and developing our optimisation-based approach, we have primarily focused our testing on static environments. Whilst we demonstrate our approach in dynamic environments, to ensure that our approach provides a robust solution, further work is required in exploring the link between the tune-able parameters and the dynamic obstacles that we expect to encounter. \revision{While it would be interesting to conduct further experiments in complex scenarios, and stress-test the approach in the presence of non-convex obstacles, these tests are more relevant for testing the motion planning method since the gaze control optimisation, as presented here, is agnostic to obstacles; we will thus explore this in future work.} Additionally, we will explore the incorporation of additional priors \revision{that may be available, such as semantic information}, into the optimisation to better observe dynamic obstacles. Humans in particular are likely to move in smooth, structured motions with a goal in mind --- a probability distribution over a known workspace could thus be used to represent the places that a human is likely to occupy -- these regions can be prioritised in the optimisation. Within the \revision{current framework, this could be achieved by assigning additional attributes to voxels for incorporation into the reward function}. Further work in this area will extend our previous work \cite{Finean2020} of using predicted composite signed distance fields to account for predicted dynamic obstacle trajectories; this framework requires the perception system to observe and track dynamic obstacles local to the robot. Accounting for the predicted movement of moving obstacles, will help us avoid scenarios in which the motion planner repeatedly plans for the robot to go into the path of the moving obstacles.

\revision{In this work, our primary focus has been to make observations that are relevant for collision-avoidance with additional exploratory reward being aimed at observing regions of space that have not been recently observed. An interesting avenue to explore in future work would be to combine this with the typical NBV problem and consider information gain in the reward function to promote efficient reconstruction of the environment. The difficulty in this approach will likely be determining the trade-off between the rewards for reconstructing the environment and for collision-avoidance, i.e. the swept trajectory term.}

A limitation that we observed when using active head behaviours is an increased error in pointcloud synchronisation and the effects of a rolling vs global shutter. Fast camera movements can result in `phantom' observations as well as inaccuracies in the position of observed objects. These errors can be mitigated by both maintaining a tight constraint on the time-synchronisation of robot joint states and received sensor measurements, as well as limiting the speed of head camera movements.

On a similar point, our method uses a weighted reward which determines the trade-off between exploring the environment and monitoring the planned trajectory ahead. These weights are likely to be determined by the environment in which the robot is operating and may evolve over time. In static environments, or scenes in which dynamic obstacles move more slowly, the time threshold can be increased.

The problem that we have addressed in this work could also merit further investigation with reinforcement learning methods, particularly for known environments where data can be accumulated over time. We believe that this may be applicable since the reward function is intuitive yet difficult to parameterise, i.e. we wish to reward better observance of the environment and behaviours that result in collision-free trajectories.

\section{Conclusion}
We provided the first description of active gaze control for trajectory-constrained mobile robots and proposed a novel solution that uses a greedy optimisation of voxelised rewards across motion primitives. We compared our method to several benchmark approaches, in 2D and 3D, and show that it outperforms in both collision avoidance and general perception of the environment. In 3D dynamic environment, we demonstrate our method to be the only approach that can reliably achieve a collision-free trajectory and avoid the moving obstacle. We further verified our findings on a physical Toyota Human Support Robot (HSR), using a GPU-accelerated perception framework, and demonstrated that our system can robustly operate in unknown and dynamic environments using live sensor measurements.

\bibliographystyle{IEEEtran}
\bibliography{main}

\end{document}